\documentclass{article}
\usepackage{spconf,amsmath,graphicx}

\usepackage{multirow}
\usepackage{booktabs}
\usepackage{makecell}
\usepackage{threeparttable}
\usepackage{makecell}
\usepackage{graphicx}

\title{Unsupervised Multimodal Image Registration with Adaptative Gradient Guidance}

\name{Zhe Xu$^{\star \dagger}$ \qquad Jiangpeng Yan$^{\star}$ \qquad Jie Luo$^{\dagger}$ \qquad Xiu Li$^{\star}$ \qquad Jayender Jagadeesan$^{\dagger}$}

\address{$^{\star}$ Shenzhen International Graduate School, Tsinghua University, Shenzhen, China \\
    $^{\dagger}$ Brigham and Women’s Hospital, Harvard Medical School, Boston, USA}

\begin{document}
%\ninept
%
\maketitle
\begin{abstract}
Multimodal image registration (MIR) is a fundamental procedure in many image-guided therapies. Recently, unsupervised learning-based methods have demonstrated promising performance over accuracy and efficiency in deformable image registration. However, the estimated deformation fields of the existing methods fully rely on the to-be-registered image pair. It is difficult for the networks to be aware of the mismatched boundaries, resulting in unsatisfactory organ boundary alignment. In this paper, we propose a novel multimodal registration framework, which leverages the deformation fields estimated from both: (i) the original to-be-registered image pair, (ii) their corresponding gradient intensity maps, and adaptively fuses them with the proposed gated fusion module. With the help of auxiliary gradient-space guidance, the network can concentrate more on the spatial relationship of the organ boundary. Experimental results on two clinically acquired CT-MRI datasets demonstrate the effectiveness of our proposed approach. 

\end{abstract}
\section{Introduction}
\label{sec:intro}
Deformable image registration (DIR) is an essential procedure in various clinical applications such as radiotherapy, image-guided interventions and preoperative planning. Recent years have witnessed that deep learning-based registration methods make magnificent progress in terms of accuracy and efficiency. In general, the learning-based methods can be categorized into the supervised \cite{lv2018supervised, hu2018weakly, hu2019dual} and the unsupervised settings \cite{VM2018, xu2020f3rnet, zhao2019unsupervised}. Among the methods, the unsupervised strategy becomes dominant recently since the supervised strategy always struggles with limited ground-truth landmarks or deformation fields. Within the unsupervised learning-based pipeline, the network can be trained under the guidance of similarity between the warped images and the fixed images. However, previous works mainly focus on unimodal DIR, multimodal DIR remains a challenging task due to non-functional intensity mapping across different modalities and limited multimodal similarity metrics. 

To generalize the applications to multimodal scenarios, some methods introduce the multimodal similarity metrics, e.g., mutual information (MI) \cite{wells1996multi} and modality independent neighborhood descriptor (MIND) \cite{mind}. However, such measures may be limited by lacking spatial prior knowledge, resulting in good global registration but relatively poor local boundary alignment. Another trend is using Generative Adversarial Network (GAN) \cite{goodfellow2014generative} to convert  the multimodal problem to the unimodal one \cite{fan2019adversarial, qin2019unsupervised, Xu2020unsupervised}. However, being a challenging task itself, the training of GAN is greatly time-consuming and hard to be controlled, and the synthetic features may lead to mismatched problems.

\begin{figure}[htb]
  \centering
   \vspace{-0.2cm}
  \centerline{\includegraphics[width=8.3cm]{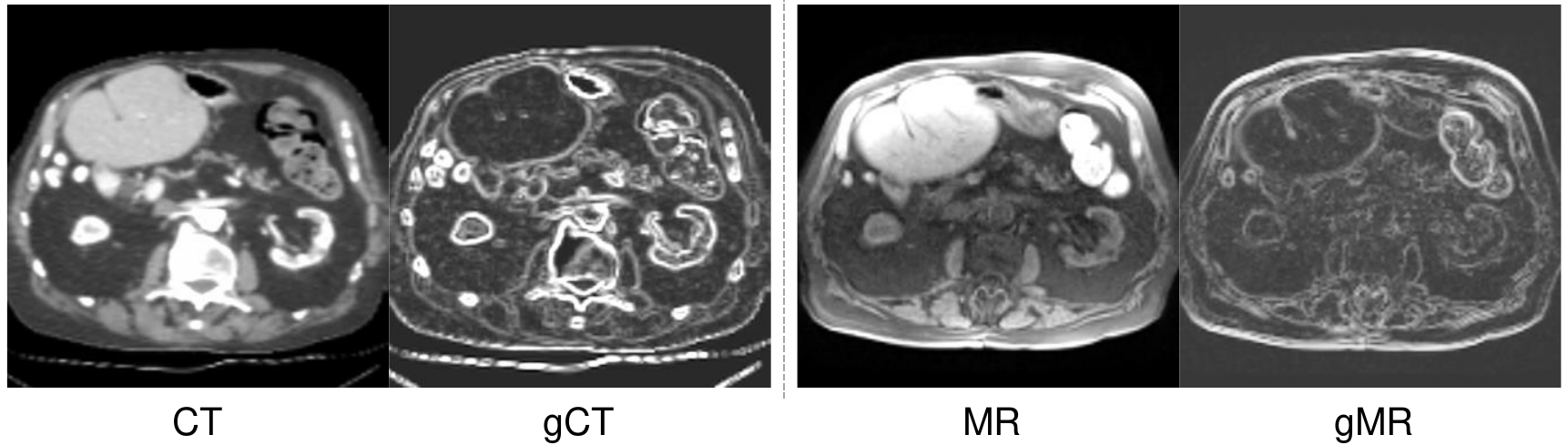}}
  \caption{Example CT and MR images and their corresponding gradient intensity maps.}
  \vspace{-0.2cm}
  \label{fig_intro}
\end{figure}

In this work, we propose a novel unsupervised multimodal image registration approach with auxiliary gradient guidance. Specifically, we exploit the gradient maps of images that highlight the biological structural boundary (shown in Fig.\ref{fig_intro}) to guide the high-quality image registration. In particular, distinct from the typical mono-branch unsupervised image registration network, our approach leverages the deformation fields estimated from two following branches: (i) image registration branch aiming to align the original moving and fixed images, (ii) gradient map registration aiming to align the corresponding gradient intensity maps. In other words, the auxiliary gradient-space registration imposes second-order supervision on the image registration. The network further adaptively fuses them via the proposed gated fusion module towards achieving the best registration performance. Contributions and advantages of our approach can be summarized as follows:

1. Our approach leverages the deformation fields estimated from the image registration branch and gradient map registration branch to concentrate more on organ boundary and achieve better registration accuracy.

2. The proposed gated dual-branch fusion module can learn how to adaptively fuse the information from two distinct branches.

Quantitative and qualitative experimental results on two clinically acquired CT-MRI datasets demonstrate the effectiveness of our proposed approach.

\section{Methods}

In this section, we first introduce the overall framework. Then we present the details of the multimodal image registration branch, the auxiliary gradient map registration branch and the gated dual-branch fusion module in Sec.\ref{framwork_describe}. Accordingly, the loss function of the framework is presented in Sec.\ref{loss} .

\subsection{Overview}
The whole pipeline of our method is depicted in Fig.\ref{fig_framework}. Given a moving $CT$ and a fixed $MR$, we calculate their corresponding gradient maps $gCT$ and $gMR$ first to obtain the auxiliary to-be-registered gradient intensity information. Then, the image registration branch estimates the primary deformation field $\phi_{i}$ for $CT$ and $MR$ while the gradient map registration branch takes $gCT$ and $gMR$ as inputs to produce the deformation field $\phi_{g}$. On top of $\phi_{i}$ and $\phi_{g}$, a gated dual-branch fusion module with output $\phi_{ig}$ is proposed to adaptively fuse the estimated deformation fields, followed by the Spatial Transformation Network (STN) \cite{STN}. It is noteworthy that the method is in a fully unsupervised manner without requiring any ground-truth deformation or segmentation label.

\subsection{Dual-Branch Image and Gradient Registration Networks}
\label{framwork_describe}
\subsubsection{Image Registration Branch}
\begin{figure}[htb]
  \centering
   \vspace{-0.2cm}
  \centerline{\includegraphics[width=9cm]{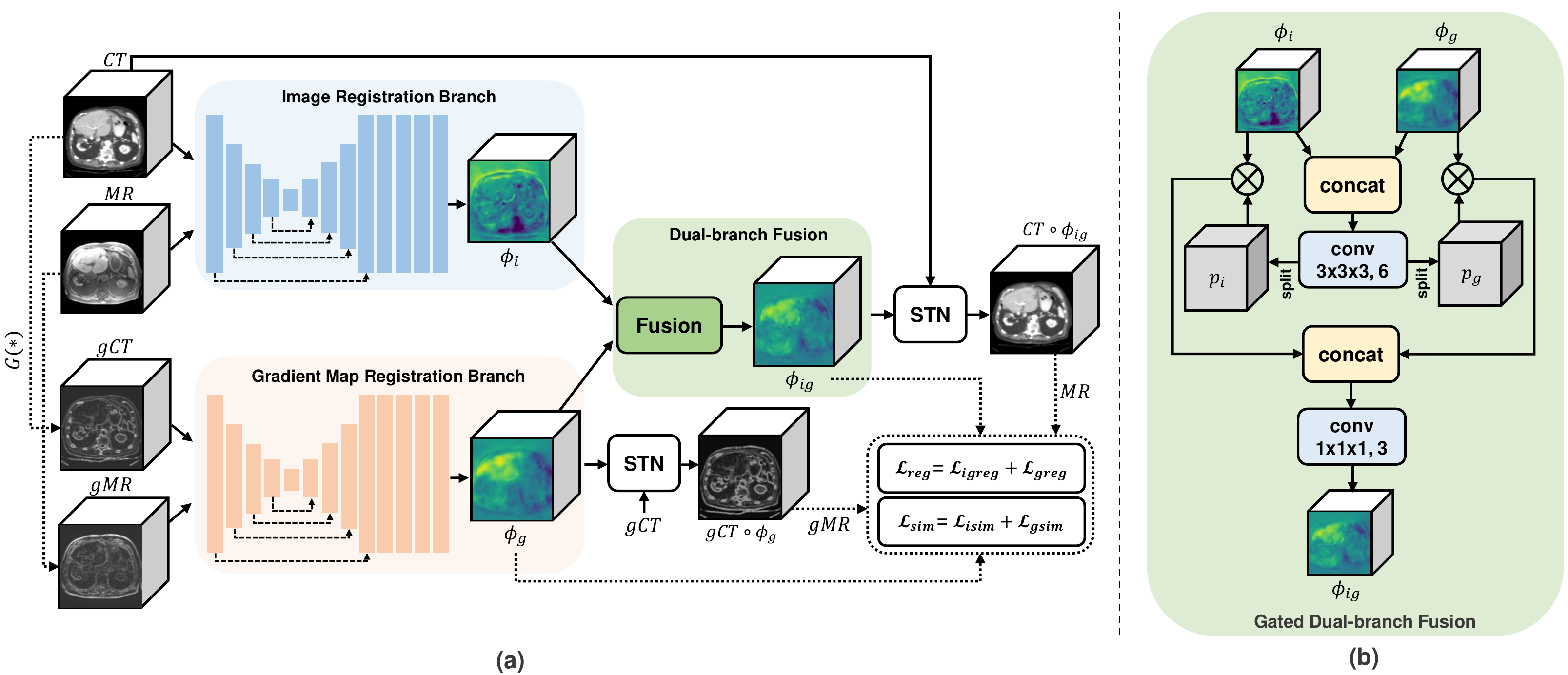}}
  \caption{Illustration of (a) our proposed framework and (b) the gated fusion module.}
  \vspace{-0.1cm}
  \label{fig_framework}
\end{figure}

The blue box in Fig.\ref{fig_framework} illustrates the general image registration branch with inputs of $CT$ and $MR$. The neural network uses the same architecture described in VoxelMorph \cite{VM2018}. Specifically, $CT$ and $MR$ are concatenated as a single 2-channel 3D image input, and downsampled by four $3 \times 3 \times 3$ convolutions with stride of 2 as the encoder. The decoder consists of several 32-filter convolutions and four upsampling operations, which can bring the image back to full resolution. Skip connections are also applied to concatenate features between encoder and decoder. Additionally, four convolutions are utilized to refine the 3-channel full-resolution deformation field $\phi_{i}$. 

\subsubsection{Gradient Map Registration Branch} 
The target of the gradient map registration branch is to estimate the auxiliary deformation $\phi_{g}$ between the moving gradient map of CT ($gCT$) and the fixed gradient map of MR ($gMR$). Specifically, the gradient map of a 3D volume $V$ can be easily obtained by computing the difference between adjacent pixels. The operation can be formulated as followed: 
\begin{equation}
\begin{aligned}
V_{x}(\mathbf{x}) &=V(x+1, y, z)-V(x-1, y, z), \\
V_{y}(\mathbf{x}) &=V(x, y+1, z)-V(x, y-1, z), \\
V_{z}(\mathbf{x}) &=V(x, y, z+1)-V(x, y, z-1), \\
\nabla V(\mathbf{x}) &=\left(V_{x}(\mathbf{x}), V_{y}(\mathbf{x}), V_{z}(\mathbf{x})\right), \\
G(V) &=\|\nabla V\|_{2},
\end{aligned}
\end{equation}
where elements of $G(V)$ are the gradient lengths for pixels with coordinates $\mathbf{x} = (x, y, z)$. In other words, we adopt the gradient intensity maps as the gradient maps without considering the gradient direction since such gradient intensity maps are sufficient to delineate the organic edge sharpness, and can be regarded as another image modality. Thus, the gradient map registration is equivalent to the spatial transformation between the CT edge sharpness and the MR edge sharpness. Furthermore, most areas of the gradient maps are close to zero, so that the subnetwork can pay more attention to the spatial relationship of the organic profiles. Therefore, the branch is more sensitive in capturing the structural dependency and can provide outline-alignment supervision to the image registration branch in turn. The magnitude of the deformation field $\phi_{g}$ can implicitly reflect whether each to-be-registered region has large or slight local deformation.

In practice, the subnetwork of this branch adopts the same architecture as in the image registration branch but takes the concatenated gradient maps as input, directly followed by an STN to apply $\phi_{g}$ to the moving $gCT$.

\subsubsection{Gated Dual-branch Fusion Module}
Since our main task is to register the original moving CT image and the fixed MR image, effectively fusing the complementary deformations $\phi_{i}$ and $\phi_{g}$ from the two different branches plays an important role in final registration. If not considered carefully, the fused deformation may otherwise yield performance degradation. As for the dual-branch deformation fusion, the most related approach \cite{wei2019synthesis} uses average operation. Hard-coding average operation considers two deformations equally, which may disregard the relative importance of each voxel from two distinct streams.  
Therefore, apart from using hard-coding operation, we further investigate a gated dual-branch fusion mechanism to automatically learn how to adaptively fuse the two distinct deformations. As shown in Fig.\ref{fig_framework} (b), gated fusion firstly uses a 6-channel 3D convolution with the sigmoid activation, which can obtain the 6-channel gating attention weight matrix ranging from 0 to 1. Then, the matrix is split into two separate gating weight maps $p_{i}$ and $p_{g}$. Next, two separate element-wise multiplications are imposed, which re-weight the $\phi_{i}$ and $\phi_{g}$ as  $\tilde{\phi}_{i}=p_{i} \cdot \phi_{i}$ and $\tilde{\phi}_{g}=p_{g} \cdot \phi_{g}$. Then, $\tilde{\phi}_{i}$ and $\tilde{\phi}_{g}$ are concatenated and fed to a bottleneck of $1 \times 1 \times 1$ convolution with 3 channels, where the final deformation field $\phi_{ig}$ is obtained.

\subsection{Loss Function}
\label{loss}
The loss function of the previous mono-branch registration networks generally includes two terms: i) the (dis)similarity between the warped image and the fixed image, ii) the regularization term for the deformation field. Distinct from the mono-branch networks, the loss function of our proposed dual-branch framework consists of four parts:
\begin{equation}
\label{loss_total}
\mathcal{L}_{total}=\left(\mathcal{L}_{isim}+ \alpha \mathcal{L}_{gsim}\right)+\left(\gamma\mathcal{L}_{igreg}+ \beta\mathcal{L}_{greg}\right),
\end{equation}
where $\mathcal{L}_{isim}$ represents the (dis)similarity between the warped CT image and the fixed MR image, while $\mathcal{L}_{gsim}$ is to measure that between the warped gCT and gMR. Likewise, the regularization terms $\mathcal{L}_{igreg}$ and $\mathcal{L}_{greg}$ are for the deformation fields $\phi_{ig}$ and $\phi_{g}$ respectively. $\alpha$, $\beta$ and $\gamma$ indicate the relative importance of $\mathcal{L}_{gsim}$, $\mathcal{L}_{greg}$ and $\mathcal{L}_{igreg}$. In the experiments, we adopt the L2-norm of gradients of deformation fields as shown below:
\begin{equation}\left\{\begin{array}{l}
\mathcal{L}_{igreg}(\phi_{ig})=\sum_{x \in \Omega}\|\nabla \phi_{ig}(x)\|_{2}; \\
\mathcal{L}_{greg}(\phi_{g})=\sum_{x \in \Omega}\|\nabla \phi_{g}(x)\|_{2}.
\end{array}\right.\end{equation}

The similarity metrics for multimodal image registration are not as flexible as in unimodal tasks. Among the limited selections of multimodal metrics, Modality Independent Neighborhood Descriptor (MIND) \cite{mind} has demonstrated its promising performance for describing the invariant structural representations across various modalities. Specifically, MIND is defined as follows:
\begin{equation}
{M}_{x}^{r}(I)=\exp \left(\frac{-D_{p}(I, x, x+r)}{V(I, x)}\right),
\end{equation}
where $M$ represents MIND features, $I$ is an image, $x$ is a location in the image, $r$ is a distance vector, $V(I, x)$ is an estimate of the local variance, and $D_{p}(I, x, x+r)$ denotes the L2 distance between two image patches $p$ respectively. 

Given the warped CT image ($CT \circ \phi_{ig}$), fixed MR image and their corresponding gradient maps $gCT$ and $gMR$, we minimize the difference of their MIND features:
\begin{equation}\left\{\begin{array}{l}
\mathcal{L}_{isim}\left(CT \circ \phi_{ig}, MR\right)\\=\frac{1}{N|R|} \sum_{x}\left\|M\left(CT \circ \phi_{ig} \right)-M\left(MR\right)\right\|_{1}; \\
\mathcal{L}_{gsim}\left(gCT \circ \phi_{g}, gMR\right)\\=\frac{1}{N|R|} \sum_{x}\left\|M\left(gCT \circ \phi_{g} \right)-M\left(gMR\right)\right\|_{1},
\end{array}\right.\end{equation}
where $N$ denotes the number of voxels in input images, $R$ is a non-local region around voxel $x$. 

\section{Experiments}
\subsection{Implementation Details}
\subsubsection{Datasets} 
Under the IRB approved study, we obtained two intra-subject CT-MR datasets containing paired CT and MR images. \\

\noindent1) \textsl{Pig Ex-vivo Kidney Dataset}. This dataset contains 18 pairs of ex-vivo kidney CT-MR scans with segmentation labels for evaluation. We carried out the standard preprocessing steps, including resampling, spatial normalization and cropping for each scan. The images were processed to $144\times80\times256$ subvolumes with 1mm isotropic voxels and divided into two groups for training (13 cases) and testing (5 cases).\\
\noindent2) \textsl{Abdomen Dataset}. This intra-patient CT-MR dataset containing 50 pairs was collected from a local hospital and annotated with the segmentation of liver, kidney and spleen. Similarly, the 3D images were preprocessed by the standard steps and cropped into $144\times144\times128$ subvolumes with the same resolution (${1mm}^{3}$). The dataset was divided into two groups for training (40 cases) and testing (10 cases).

\subsubsection{Training Strategies}
The proposed framework was implemented on Keras with the Tensorflow backend and trained on an NVIDIA Titan X (Pascal) GPU. For both datasets, we adopted Adam as optimizer with a learning rate of 1e-5, and set the batch size to 1. $\alpha$, $\beta$ and $\gamma$ were set to 0.4, 0.3, 0.8 for pig kidney dataset, and 0.5, 0.5, 1 for abdomen dataset through grid searching.

\subsection{Experimental Results}
\subsubsection{Quantitative Comparison}
Within our framework, two fusion mechanisms (denoted as Ours(average) and Ours(gated)) were adopted for comparison. In addition, we compared our approaches with the top-performing conventional method SyN \cite{Avants2008SymmetricDI} and two VoxelMorph-based \cite{VM2018} mono-branch unsupervised networks  (denoted as VM and VM(concat)) with MIND-based similarity metric. Specifically, VM concatenates moving image and fixed image as a 2-channel input, while VM(concat) additionally concatenates their gradient maps as a 4-channel input. 

We quantitatively evaluated our method over two criteria: Dice score of the organ segmentation masks (DSC) and the average surface distance (ASD). The results are presented in Table \ref{tab_result}. For the pig ex-vivo kidney dataset, the intensity distributions of the ex-vivo organ scan are simpler. It can be seen that the two variants of our methods both slightly outperform other baselines in terms of DSC and the performances of the two fusion mechanisms are close. Turning to the more complex abdomen dataset, our approaches both significantly achieve higher DSC and lower ASD than the conventional and learning-based approaches. Not surprisingly, VM(concat) shows slight improvement compared to VM, which demonstrates that the gradient-space information can provide useful information for image alignment. With the dual-branch learning fashion, image registration can further benefit from the gradient information. In particular, the gated fusion mechanism obtains more improvements in most organ alignments than the average fusion. Furthermore, the conventional approach SyN costs more than 5 minutes to estimate a transformation, while all the learning-based methods only cost less than a second with a GPU.

\begin{table}[]
\centering
\caption{Comparison of DSC and ASD on different methods. The best and the second results are shown in bold and underline respectively.}\label{tab_result}
\scalebox{0.71}{
\begin{tabular}{c|c|ccc|ccc}
\Xhline{1pt}
                        \multicolumn{1}{c|}{Dataset}                      & \multicolumn{1}{c|}{Pig}                              & \multicolumn{6}{c}{Abdomen}                                                                                                                                            \\ \hline
\multicolumn{1}{c|}{\multirow{2}{*}{Method}}  & \multicolumn{1}{c|}{Dice(\%)} & \multicolumn{3}{c|}{Dice(\%)}                                                           & \multicolumn{3}{c}{ASD(mm)}                                                          \\ \cline{2-8} 
\multicolumn{1}{c|}{}                        & \multicolumn{1}{c|}{Kidney}  & \multicolumn{1}{c}{Liver}   & \multicolumn{1}{c}{Spleen} & \multicolumn{1}{c|}{Kidney} & \multicolumn{1}{c}{Liver} & \multicolumn{1}{c}{Spleen} & \multicolumn{1}{c}{Kidney}  \\ \hline
Moving                                        &       84.29                                                      &         77.18                   &         78.24                    &              80.14               &          4.95                  &         1.97                    &   2.01                          \\ 
SyN(MI)                                       &        85.17 & 79.18                                                     &        80.21                                                 &        82.91                     &     4.81                       &      1.54                       &  1.92                           \\ 
VM                                         &      91.24 & 84.17                                                      &        82.76                    &         83.51                    &           3.92                  &          1.47                  &           1.72                              \\ 
VM(concat)                           &   91.51    &                  85.72                                     &     84.23                       &             83.96                &          3.03                  &         \underline{1.27}             &                  1.44                                   \\ \hline
Ours(average)                                 &   \underline{92.09}                                                        &     \underline{86.25}                       &          \underline{85.03}                   &         \underline{84.39}                    &        \underline{2.94}                    &        \textbf{1.22}                     &    \underline{1.37}                            \\ 
Ours(gated)                                   &      \textbf{92.13}                                                        &      \textbf{87.66}                       &           \textbf{86.83}                   &          \textbf{85.07}                    &         \textbf{2.73}                    &        1.31                     &     \textbf{1.25}                         \\ \Xhline{1pt}
\end{tabular}}
\end{table}

\subsubsection{Qualitative Comparison}
\begin{figure}[htb]
  \centering
%   \vspace{-0.35cm}
  \centerline{\includegraphics[width=8.2cm]{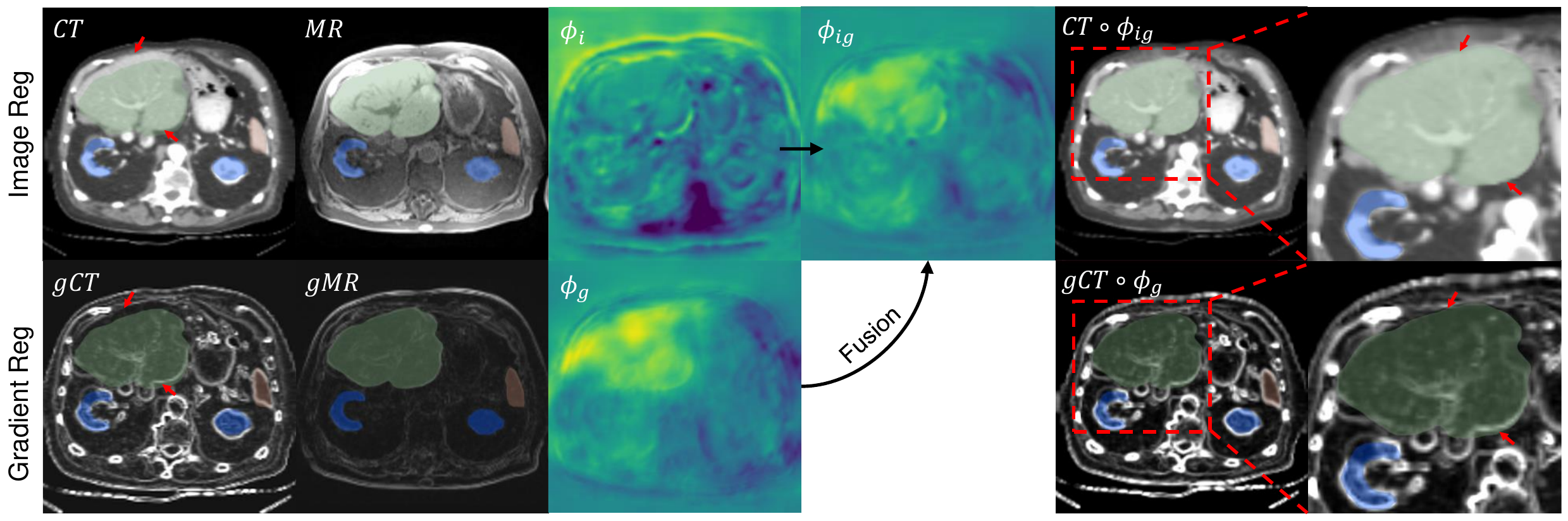}}
  \caption{Example registration results within our framework.}
  \vspace{-0.3cm}
  \label{fig_inside}
\end{figure}

Fig.\ref{fig_inside} visualizes a registration example of the abdomen dataset within our framework using the gated fusion module. In abdominal scans, livers with tumors usually have large local deformation due to progressed disease, patient motion and insufflation during surgery, while the deformations of surrounding organs are less obvious.

Given the original to-be-registered images and their corresponding gradient maps, the dual-branch networks can not only effectively register the CT image to the MR image, but also align the gradient intensity maps in the gradient map registration branch. The zoom-in images show that the liver and its gradient outline are more accurately aligned.

\begin{figure}[htb]
  \centering
  \centerline{\includegraphics[width=8.5cm]{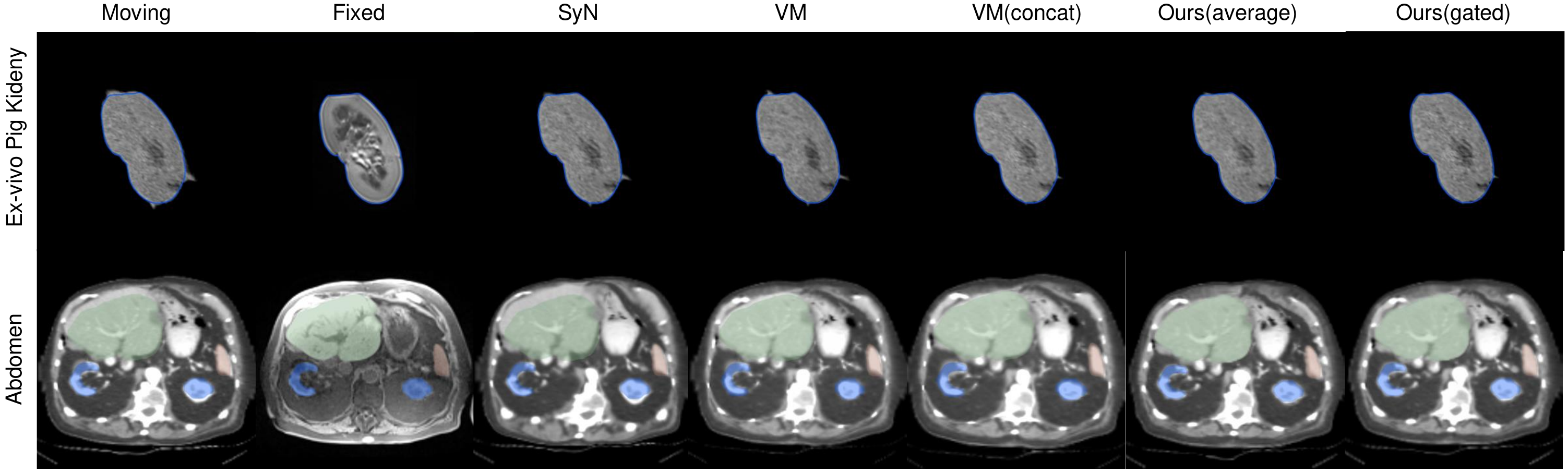}}
  \caption{Qualitative comparison with other methods. The segmentation of the organs are depicted as outlines (in pig kidney dataset) and masks (in abdomen dataset).}
  \vspace{-0.1cm}
  \label{fig_comparison}
\end{figure}
Following the example in Fig.\ref{fig_inside}, we visually compare our proposed method with other aforementioned baselines as shown in Fig.\ref{fig_comparison}. Consistent with the quantitative results, our approaches achieve more accurate organ boundary alignments, especially in the abdominal case. The visual comparison further proves that our proposed auxiliary gradient branch helps the network be more sensitive to the hard-to-aligned local regions by considering geometric organ structures.

\section{Conclusion}
In this paper, we propose a novel unsupervised multimodal image registration method with auxiliary gradient guidance to further improve the performance of organ boundary alignment. Distinct from the typical mono-branch unsupervised image registration network, our approach leverages not only the original to-be-registered images but also their corresponding gradient intensity maps in a dual-branch adaptative registration fashion. Quantitative and qualitative results on two clinically acquired CT-MR datasets demonstrate the effectiveness of our proposed approach. 

\section{Compliance with Ethical Standards}
All procedures performed in studies involving human participants were in accordance with the ethical standards of the institutional and/or national research committee and with the 1964 Helsinki Declaration and its later amendments or comparable ethical standards.

\section{Acknowledgements}
This project was supported by the National Institutes of Health (Grant No. R01EB025964, R01DK119269, and P41EB015898), the National Key R\&D Program of China (No. 2020AAA0108303), NSFC 41876098 and the Overseas Cooperation Research Fund of Tsinghua Shenzhen International Graduate School (Grant No. HW2018008).

\bibliographystyle{IEEEbib}
\bibliography{strings,refs}

\end{document}